\documentclass[preprint]{elsarticle}

\usepackage{lineno,hyperref}
\modulolinenumbers[5]
\usepackage{xcolor}
\usepackage{comment}
\usepackage{times}
\usepackage{epsfig}
\usepackage{graphicx, subfig, longtable}
\usepackage{amsmath}
\usepackage{amssymb}
\usepackage{breqn}
\usepackage{array}
\usepackage{textcomp}
\newcolumntype{P}[1]{>{\centering\arraybackslash}p{#1}}
\usepackage{multirow}
\usepackage{ulem}
\usepackage{boldline}
\usepackage{url}
\usepackage{enumitem}
\usepackage{tabularx}
\usepackage{float}
\journal{ISPRS Journal of Photogrammetry and Remote Sensing}

\biboptions{authoryear}

\begin{document}

\begin{frontmatter}

\title{Weakly-Supervised Domain Adaptation for Built-up Region Segmentation in Aerial and Satellite Imagery}





\author{Javed~Iqbal}
\ead{javed.iqbal@itu.edu.pk}

\author{Mohsen~Ali\corref{cor}}
\cortext[cor]{Corresponding author}
\ead{mohsen.ali@itu.edu.pk}

\address{Information Technology University, Pakistan}


\begin{abstract}
This paper proposes a novel domain adaptation algorithm to  handle the challenges posed by the satellite and aerial imagery, and demonstrates its effectiveness on the built-up region segmentation problem.
Built-up area estimation is an important component in understanding the human impact on the environment, effect of public policy and in general urban population analysis.
The diverse nature of aerial and satellite imagery (capturing different geographical locations, terrains and weather conditions) and lack of labeled data covering this diversity makes machine learning algorithms difficult to generalize for such tasks, especially across multiple domains.
Re-training for new domain is both computationally and labor expansive mainly due to the cost of collecting pixel level labels required for the segmentation task.
Domain adaptation algorithms have been proposed to enable algorithms trained on images of one domain (source) to work on images from other dataset (target). Unsupervised domain adaptation is a popular choice since it allows the trained model to adapt without requiring any ground-truth information of the target domain.
On the other hand,
due to the lack of strong spatial context and structure, in comparison to the ground imagery, application of existing unsupervised domain adaptation methods results in the sub-optimal adaptation.
We thoroughly study limitations of existing domain adaptation methods and propose a weakly-supervised adaptation strategy where we assume image level labels are available for the target domain. 
More specifically, we design a built-up area segmentation network (as encoder-decoder), with image classification head added to guide the adaptation.
The devised system is able to address the problem of visual differences in multiple satellite and aerial imagery datasets, ranging from high resolution (HR) to very high resolution (VHR), by investigating the latent space as well as the structured output space.

A realistic and challenging HR dataset is created by hand-tagging the 73.4 sq-km of Rwanda, capturing a variety of build-up structures over different terrain. The developed dataset is spatially rich compared to existing datasets and covers diverse built-up scenarios including built-up areas in forests and deserts, mud houses, tin and colored rooftops. Extensive experiments are performed by adapting from the single-source domain datasets, such as Massachusetts Buildings Dataset, to segment out the target domain. We achieve high gains ranging 11.6\%-52\% in IoU over the existing state-of-the-art methods. 
\end{abstract}


\end{frontmatter}


\section{Introduction}
Built-up area estimation (\cite{chowdhury2018estimating, tan2018precise}) is an important component in understanding the human impact on the environment, public policy and urban population analysis.  
However, manual methods using survey and census are both difficult and expensive to apply. 
With satellite imagery becoming available, land-use and land-cover analysis through deep learning is increasingly gaining interest. 
Built-up area estimation and built-up area segmentation is one of the sub-problems in land-use analysis (\cite{DLinRS, landuse&landcover, Detecting-boundaries}). 
The intense variations in objects and scenes captured in satellite imagery are difficult to be learned by a single model trained on a dataset collected around few specific locations. 
These variations are more prominent when objects of interest are man made structures like buildings. In addition different weather conditions and time of the day (when image was obtained) also affect the appearance of those structures and their surrounding (\cite{facebook_building_detection}).
Therefore, a system trained for semantic segmentation of satellite imagery on a specific data will not generalize to another set which is unseen to the system during training. 

The abundantly available satellite imagery that captures diversity across the globe is unlabeled and it is difficult to develop separate deep models for new geographic locations in the absence of labeled data. 
In this work, we aim to perform single class semantic segmentation (built-up region segmentation) and then adaptation of the learned model to unseen data using only the visible spectrum, e.g., RGB (Red, Blue and Green) bands of satellite imagery. 
To handle the extreme diversity, we introduce a novel method of adapting both the latent space and the structured output space, with an assumption of image level weakly labeled data. 
Unlike the previous methods (\cite{chen2017no, benjdira2019unsupervised}) where the feature space adaption is performed, our weakly labeled based latent space adaption does not suffer from adapting the high-dimensional features. 




\begin{figure}[t]
	\centering
	\includegraphics[width=3.8in]{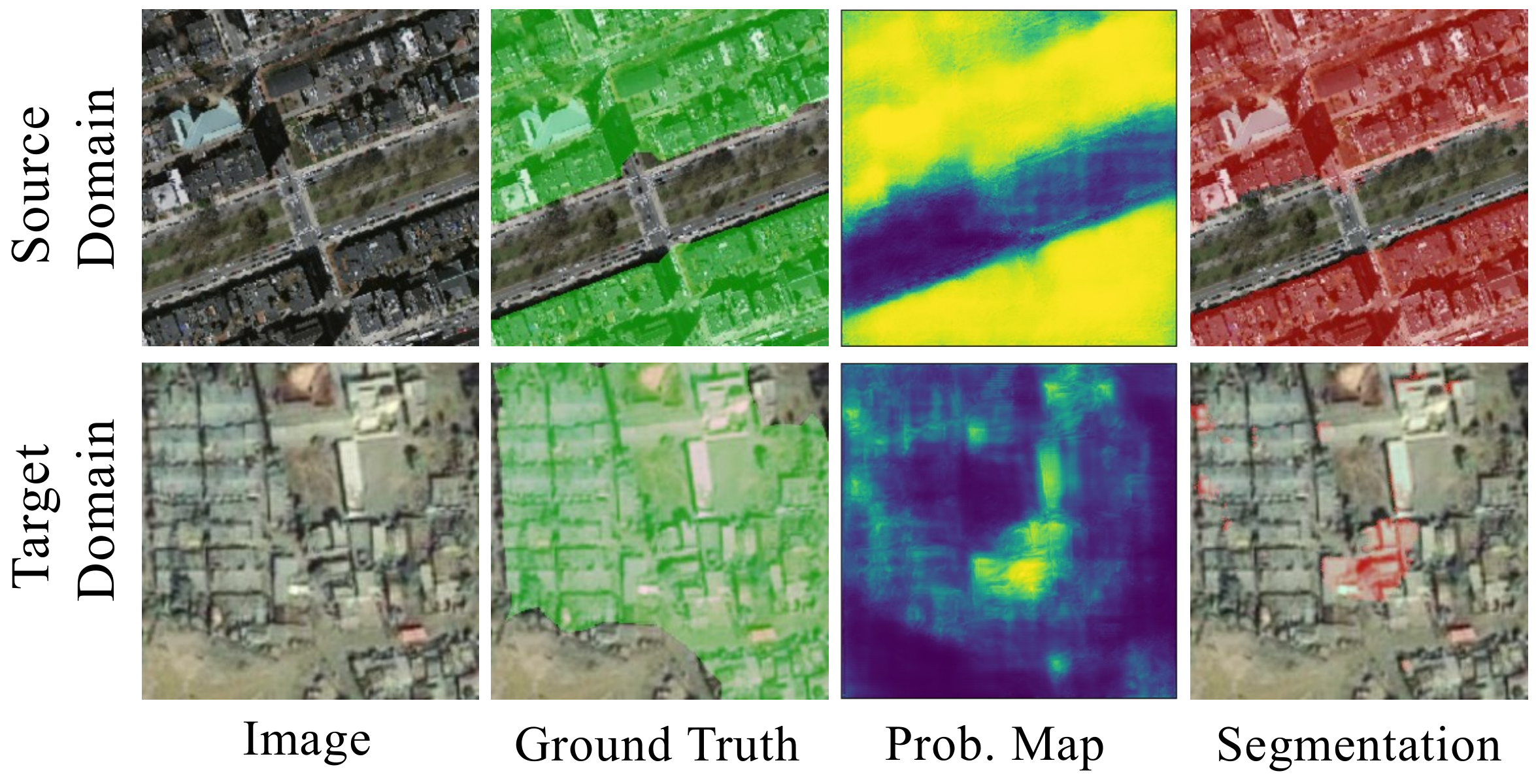}
	\caption{Built-up area segmentation results for source and target domain images. The source domain trained model fails to produce effective results in the target domain (best viewed in color).}
	\label{img:0}
\end{figure}

Semantic segmentation of satellite imagery suffers from similar challenges as in ground imagery, e.g., difficulty of generalization to unseen data, unavailability of the labeled data and complexity in creating a labeled data. 
However, satellite imagery does differ from the ground image datasets such as Pascal VOC2012 (\cite{pascal-voc-2012}), Microsoft COCO (\cite{lin2014microsoft}), and Cityscapes (\cite{Cordts2016Cityscapes}) which contain everyday objects and road driving scenes respectively.
These objects have typical formation and mostly exist in specific scenarios and context, e.g., a personal computer is most likely to be found resting on a table and a car has high probability to be present on road or parked in a garage or parking lot. 
These objects mostly differ only in terms of size, color and visual appearance.
Most of the domain adaption works show their results on the similar scenes (driving scenes), with high similarity of structure and context in between the domains (both road scenes) and content (roads in the middle of the scene, sky is on the top part) (\cite{chen2017road, tsai2018learning, chen2017no, mancini2018boosting,zhang2018fully, sankaranarayanan2018learning}). 
No such structure or context exists in the case of built-up area segmentation from the satellite imagery or aerial imagery. 
Buildings, being man-made, exist in a variety of architectures, colors and sizes with regular and irregular context information depending on the cultural and economic development of a geographic location. 
Further, satellite imagery seems to be non-identical relative to different weather conditions, different terrain and more importantly due to differences in various sensors used to photograph different geographical regions at different times.
These differences make it difficult for current state-of-the-art segmentation and domain adaptation networks to generalize from ground imagery to satellite imagery. 

\begin{figure}[t]
	\centering
	\includegraphics[width=4.8in]{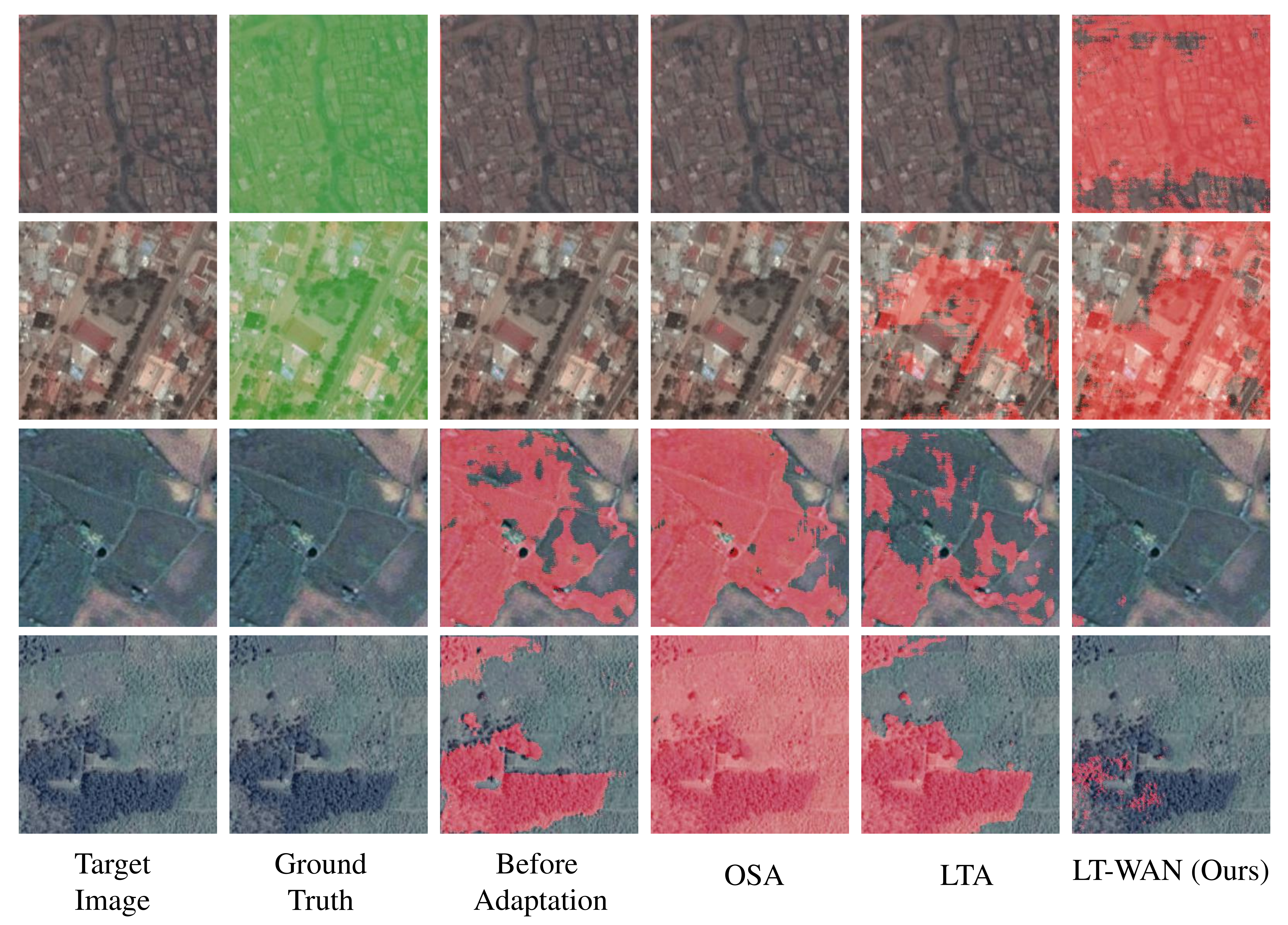}
	\caption{Domain adaptation results of existing state-of-the-art methods compared to the proposed Weakly-supervised Adaptation Network (LT-WAN). OSA: Output Space Adaptation, LTA: Latent Space Adaptation (best viewed in color).}
	\label{img:critic}
\end{figure}

Domain adaptation of built-up area segmentation is a complex task and requires to address two main problems. The false detection (no built-up regions are present in actual) and the miss-detection (built-up regions present in actual) should be minimized.
As shown in Fig. \ref{img:critic}, the current state-of-the-art domain adaptation methods (\cite{tsai2018learning, chen2017no, benjdira2019unsupervised}) are not capable to effectively adapt to unseen satellite imagery datasets. Especially, the houses built in deserts and forest in the target domain are hard to segment and hence difficult to adapt. 
Many existing methods extensively rely on adversarial loss to adapt to target datasets, which also inherits the base segmentation model limitations making the adaptation process less effective. 
 
In this work the domain adaptation for semantic segmentation is studied and investigated thoroughly. 
A weakly-supervised domain adaptation network based on Generative Adversarial Networks (GAN) (\cite{goodfellow2014generative}) and a built-up area detection module is proposed for satellite and aerial imagery. 
The GAN minimizes the distribution gap between source and target domain at output and latent space. 
The built-up area detection module using weak-labels (presence or absence of built-up areas in an image) extracts features of built-up areas to help the adaptation process.
To evaluate and test the proposed method over a variety of different scenarios, a new satellite imagery dataset for built-up regions segmentation is created. 
To the best of our knowledge, mostly available aerial and satellite imagery datasets belong to few developed locations, confined to a specific landscape with similar structural design and uniform weather conditions. 
ISPRS-Potsdam 2D semantic labeling dataset (\cite{ISPRSD}) contains aerial imagery only of Potsdam  and so is the Massachusetts buildings datasets (\cite{MnihThesis}).
On the other hand the proposed dataset is collected from Rwanda to capture intra-class variations. 
It contains a variety of differently designed built-structures with varying landscape and different weather conditions. 
Being a developing country it contains built-up regions spread irregularly in deep forests and deserts, which makes this data challenging and difficult to test on.
The main contributions of this work are as follows:
\begin{itemize}
  \item We propose a weakly-supervised domain adaptation for built-up region segmentation by leveraging the image level weakly-labeled data to directly penalize the encoder and decoder of the segmentation model. 
  This module adds significantly to the performance of the domain adaptation methods. 
  \item We exploit the domain adaptation process for aerial and satellite imagery in latent space and output space augmented with weakly-supervised built-up area detection.
  \item We developed Rwanda built-up regions dataset capturing characteristics like mud houses, tin and colored rooftops, irregular built-up areas in forests and deserts, etc., to test the robustness of the proposed algorithm.
\end{itemize}

The built-up area detection module ensures high true segmentation rate and minimizes the false positive pixel classification rate significantly. The output space domain adaptation is less effective due to absence of structured output in satellite imagery compared to latent space domain adaptation as shown in Table \ref{table:2}. The proposed approach outperforms the baseline and other state-of-the-art methods on several benchmarks datasets. 
\begin{figure*}[t]
	\centering
	\includegraphics[width=4.8in]{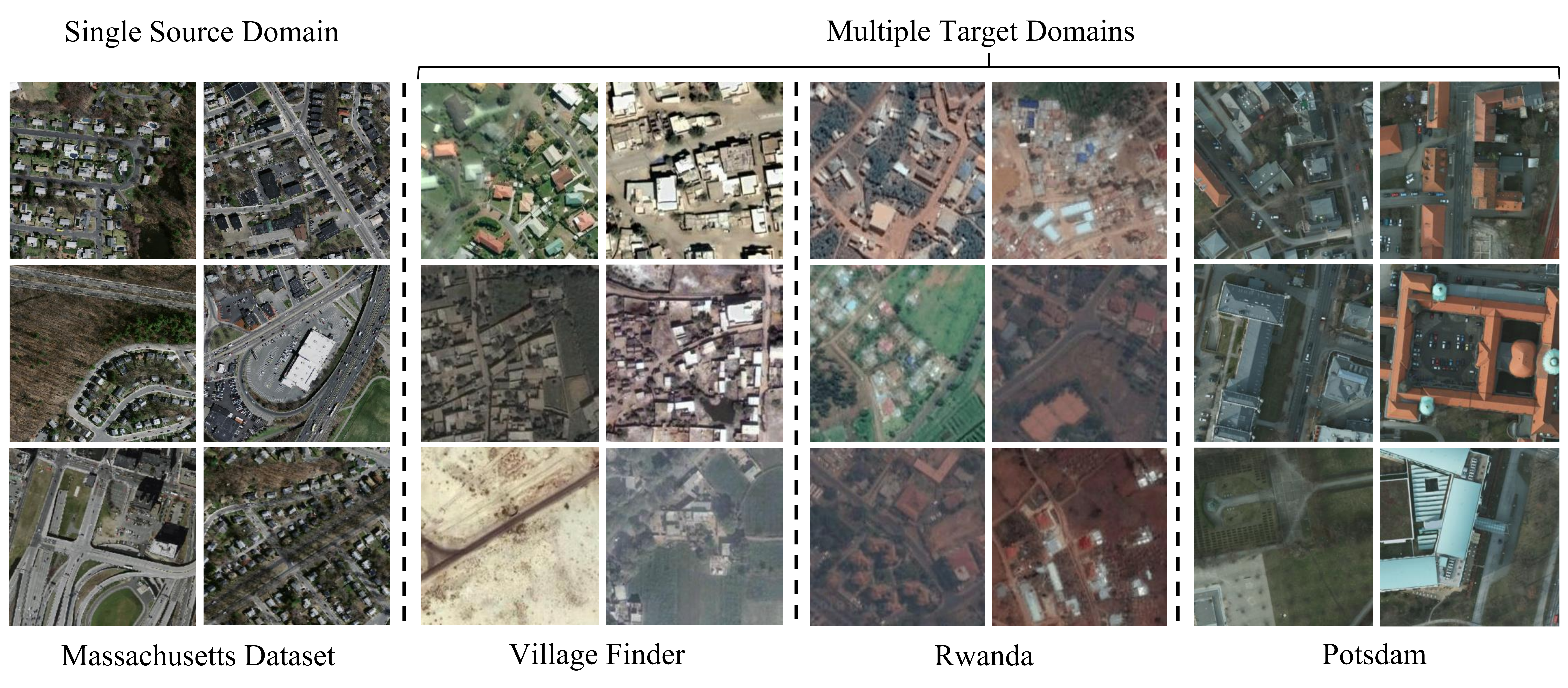}
	\caption{The figure displays sample images from both source and target domain datasets. Built-structures in source and all three target datasets are different from each other. 
	(best viewed in color)}
	\label{img:2}
\end{figure*}


\section{Related Work}
In semantic segmentation an individual label is assigned to each pixel in an image compared to image level labeling where a single label is assigned to a whole image (\cite{csurka2008simple}). 
Unlike instance based segmentation (\cite{he2017mask}), semantic segmentation does not differ between two instances of the same object.
The need of pixel-level ground-truth data makes deep Learning based semantic segmentation to be one of the important candidates for domain adaptation.
Below, we briefly review existing techniques of semantic segmentation and domain adaptation and highlight their differences and limitations. 

\textbf{Segmentation:} To address the problem of spatial resolution for semantic segmentation, many algorithms have been devised, mainly consisting of three steps i.e. convolutions, downsampling and upsampling. \cite{long2015fully} proposed a fully convolutional network (FCN) for semantic segmentation. Their proposed end-to-end FCN was based on the imagenet pre-trained models. They also introduced the concept of upsampling using deconvolution. SegNet (\cite{badrinarayanan2015segnet}) improved FCN by transferring the maxpooling indices to the upsampling side (decoder) to enhance the segmentation results and minimize memory consumption. 
DeconvNet (\cite{noh2015learning}) presented an unpooling and deep deconvolution based decoder instead of a simple upsampling process to overcome the limitation of prior algorithms. During a biomedical image segmentation challenge ISBI-2012, \cite{ronneberger2015u} proposed U-Net with a special property of skip connections between encoder and decoder layers. 
However, these methods were failing to maintain the spatial resolution by making the receptive field larger using pooling operations.
DeepLab (\cite{chen2014semantic, chen2018deeplab}) proposed dilated convolutions in FCN to address the issue of spatial resolution and introduced fully connected conditional random fields (CRF) as a smoothing operation for semantic segmentation.

Success of deep learning based semantic segmentation led it to be used on aerial and satellite imagery, localizing and segmenting the roads, vegetation, buildings, etc.
\cite{road_building_cnn} used FCN to predict roads and buildings in an aerial imagery dataset. 
\cite{ACCV_segmentation} used false color imagery with vegetation index and digital surface models (DSM) to train an encoder-decoder based architecture that performs segmentation.
Similarly, SegNet based multi-scale FCN was trained by (\cite{beyond_rgb}  to semantically segment the satellite imagery. 
\cite{Hazirbas2016FuseNetID} investigated the early and late fusion of remote sensing to integrate the information of depth in semantic segmentation.
A recent work, \cite{treeSegNet} automatically constructs adaptive tree-shaped CNN blocks for semantic segmentation. 
To cater the scale variations of objects in satellite imagery, multi-scale models are used.
\cite{Chen2014VehicleDI} trained CNN on multiple scales in order to detect vehicles, whereas (\cite{multi-scale_classification}) used an ensemble of CNN on a pyramid of images. 
Prior to deep learning based approaches, \cite{xiaogang2013extracting} leveraged impervious surface extraction with the help of object oriented methods. They augmented gravity models to estimate the gravity of areas and grade the extracted areas as built-up regions. \cite{yuksel2012automated} used the shadow and its direction to extract buildings from satellite imagery. Similarly, \cite{sirmacek2010probabilistic} defined a probabilistic framework to detect building in aerial and satellite imagery. Initially, they extracted multiple local features and then used variable-kernel density estimation to estimate the  corresponding probability density functions. However, the success of deep features based approaches mostly replaced these methods.

Built-up area segmentation differs from the building detection in a way that roads, playgrounds, parks, vegetation, or empty spaces surrounded by the buildings are also assumed to be part of built-up area (\cite{office20132011,li2015cauchy,murtaza2009villagefinder}). For our problem, we do not differentiate between rural or urban areas. 
\cite{sirmacek2009urban} applied graph matching algorithm to find the built-up area (they use the term urban area), where graph nodes were created on the basis of SIFT keypoints.
\cite{yu2018urban} detected the built-up area using the nighttime light (NTL) composite data. \cite{hu2016representation} computed Harris Features Points (HFP) as well other spectral, textural, and structural features at multiple scales by dividing the image into small blocks. Village-Finder (\cite{murtaza2009villagefinder}) used frequency and color features to train weak-classifiers which are then combined through Adaboost. \cite{bramhe2018extraction} fine-tuned the VGG (\cite{simonyan2014very}) and Inception-V3 (\cite{szegedy2016rethinking}) over the sentinel-2 from the satellite imagery for the extraction of built-up areas in sentinel-2 imagery. Similarly, (\cite{tian2018urban}) employed a strategy similar to bag-of-visual words, where the visual dictionary is built by extracting features from deep convolutional neural network (CNN). Nearest word criterion is used to classify the test images. The authors in (\cite{zhang2018built}) used CNN to extract built-up area from landsat-8 imagery.

In order to overcome the computational and annotation cost of training a fully supervised semantic segmentation model, many researchers used image level labels for weakly-supervised semantic segmentation (\cite{zhou2016learning_cam, kolesnikov2016seed, khoreva2017simple, ahn2019weakly, ali2020destruction}). \cite{zhou2016learning_cam} used global average pooling (GAP) to pool the features and trained an image classification network over it. Based on the classifier output, they obtained localization by defining class activation maps (CAM). \cite{kolesnikov2016seed} used a CAM based approach to generate localization cues based on a strong classification network. These localization cues were then expanded and contracted using a novel loss function. Similarly, \cite{khoreva2017simple} used bounding box annotations as weak labels during training and obtained better results with more refined boundaries compared to previous methods.  
\cite{ali2020destruction} employed a weakly-supervised approach with sparsity constraints, hard negative mining and conditional random fields for destruction detection in satellite imagery. Their proposed attention based mechanism is able to differentiate between image patches with destruction from normal. \cite{shakeel2019deep} used a regression model to count built-structures in satellite imagery. They defined patch level house counts to predict the respective built-structure density.  
Most of the weakly-supervised methods present in literature are mainly defined for ground imagery, but it can be applied to satellite and aerial imagery.

All previous works involved training and testing data captured from specific locations and so the models need to be fine-tuned or fully-trained to accurately localize objects in a different scenario, i.e., a different geographic location. 
The work of \cite{facebook_building_detection} evoked such issues and trained a global model that can be deployed on any location.
Differences between the training and testing resolution and zoom levels (much different from detecting objects at different scales) also affect the accuracy of models.

\textbf{Domain Adaptation:} Semantic segmentation algorithms fail to generalize and segment poorly when presented with images from domains other than where the algorithms were trained. 
This occurs due to the changes observed in the properties of sensors used or difference of object's appearance, (for example built structures, roads and other surroundings) and is named as domain shift (\cite{tsai2018learning}).
Multiple ways are adopted to solve this domain shift problem, i.e., fine tuning the pre-trained network (trained on source dataset) to the new domain (target) dataset or training on composite dataset which is not always realistic because there is a lot of annotated data required in the target domain which is not available. 
To handle this problem in an unsupervised manner, algorithms try to adapt either at the feature space or the output-space.
\cite{chen2017no} proposed an adversarial learning based domain adaptation of the feature space for the task of semantic segmentation.
They combined global domain adaptation with category level adaptation and found that the composition provides better improvement rates in the new domain. 
The authors in (\cite{mancini2018boosting}) investigated the latent space to find out the existing data distributions in the target space and adopt it to the trained network in a supervised manner. 
Similarly, \cite{chen2017road} also devised a domain adaptation strategy for road reality oriented adaptation in urban scenes. 
On the other end, (\cite{tsai2018learning}) proposed structured output domain adaptation using GAN. 
Their method tried to adapt the target domain from structured output by minimizing adversarial loss.
FCAN (\cite{zhang2018fully}) proposed an unsupervised adversarial learning based appearance adaptation augmented with representation adaptation system for domain adaptation of semantic segmentation and presented state-of-the-art results.
\cite{rebuffi2017learning} proposed domain adaptation modules based on residual blocks for a fully supervised system while preserving the performance in original domain. Based on this concept, \cite{marsde2018people} proposed a fully supervised domain adaptation to multiple target domains simultaneously with the help of domain specific adaptation modules. They specifically target the counting task and adopt a single architecture to multiple domains. The authors in (\cite{ghassemi2019learning}) extended this concept of domain specific batch normalization to aerial imagery segmentation.
Similar to methods for domain adaptation for ground imagery (\cite{chen2017no, tsai2018learning}), \cite{benjdira2019unsupervised} used GAN based feature space domain adaptation for semantic segmentation of aerial images for building detection problem. \cite{tasar2020colormapgan} proposed ColorMapGAN, a GAN based domain adaptation approach with image-to-image translation method for source to target domain images. Specifically, they shifted the source domain images to look like target domain images and then used their labels to train semantic segmentation models.
However, all the current state-of-the-art methods are suffering from absence of object label information and only exploits the semantics and dataset distributions, which are insufficient to capture the highly unstructured satellite and aerial imagery.

Besides all the existing methods, our proposed method incorporates the weak-labels (image-level labels) and exploits the latent space and output space for domain adaptation. This introduces an attention effect for built-up regions in the domain adaptation process especially where there are houses in forests, deserts and uncommon built-ups.

\section{Domain Adaptation for Satellite Imagery}
In this section the proposed domain adaptation technique for built-up region segmentation in satellite imagery is described in detail. 
We start with the basic explanation of how the popular domain adaptation methods work, discuss their limitations and then build our solution over it.
To build a baseline, we start with the existing state-of-the-art methods (\cite{tsai2018learning,chen2017no}) for unsupervised domain adaptation (UDA) for semantic segmentation, which are based on GAN (\cite{goodfellow2014generative}). In general, there are two complementary modules in a GAN-based domain adaptation approach; e.g., generator module (network) to generate representation or segmentation output and discriminator module (network) to differentiate between source and target domain datasets.

\subsection{Preliminaries}
Let $I_s \in \mathbb{R} ^{H\times W\times 3}$ be the source domain RGB images with corresponding labels $Y_s \in \mathbb{Z}_2 ^{H\times W\times C}$, where $C$ represents number of classes. Similarly for the target domain, $I_t \in \mathbb{R} ^{H\times W\times 3}$ represents RGB images with no associated labels available. 
The generator network (segmentation network) denoted as $G$ is trained using $I_s$ and $Y_s$ of the source domain. For target domain images $I_t$, the latent space representation $P_{L_t}\in \mathbb{R} ^{h\times w\times c}$ or the output space segmentation $P_{O_t} \in \mathbb{R} ^{H\times W\times C}$ is obtained using $G$, where $h, w$ and $c$ represents the height, width and depth of the latent space representation and $\sum_{C} P_{O_t} (H, W) = 1$. A global domain discriminator denoted as $D$ is defined to produce adversarial loss by differentiating between target and source domain representations (latent/output space). 
The initial objective function is adopted from (\cite{tsai2018learning}) having a weighted generator and discriminator loss factors as shown in  Eq. \ref{eqn:1}.
\begin{equation}
L{(I_s, Y_s, I_t)} = L_{seg} (I_s, Y_s) + \lambda_{adv} L_{adv} (I_t)
\label{eqn:1}
\end{equation}
where $L_{seg} (I_s, Y_s)$ is the segmentation loss of the source domain and $L_{adv} (I_t)$ is the adversarial loss of the target domain, whereas $\lambda_{adv}$ is the scaling factor to adversarial loss. In this UDA approach, the optimization function is given by the following min-max criterion:
\begin{equation}
\underset{G}{min}  ~\underset{D}{max}  ~L{(I_s, Y_s,  I_t)}
\label{eqn:2}
\end{equation}

\subsection{Adversarial Learning Based UDA for Satellite Imagery}
\label{sec:uda}
If the ground truth labels for target domain images are there, a simple idea would be to fine-tune $G$ with that available labeled data. However, it is difficult to label the huge incoming satellite imagery and then update $G$. The UDA approaches try to adapt the unlabeled target data, with labels available only for source data. For a common UDA approach for satellite imagery segmentation, the complementary sub-networks are discussed below.

\textbf{Generator:} The generator network $G$ is the U-Net (\cite{ronneberger2015u}) segmentation network. In this work, the segmentation network is a pixel level binary classifier for built-up region segmentation trained in a fully supervised way on a source dataset using the binary cross entropy loss defined in Eq. \ref{eqn:3}. 
\begin{equation}
\begin{split}
    L_{seg} (I_s, Y_s) = -\sum_{i=1}^H \sum_{j=1}^W [Y_s(i, j) log(P_{O_s}) +  (1 - Y_s(i, j))(1 - log(P_{O_s}))]
\end{split}
\label{eqn:3}
\end{equation}
where $P_{O_s}$ is the binary segmentation output of source images and $P_{O_s} = G(I_s) \in \mathbb{R} ^{H\times W\times 1}$. 

\textbf{Discriminator:} In our UDA model, the discriminator network $D$ is a binary classifier, distinguishing between source domain and target domain representations.
Using the segmentation network output denoted as $P_{O}$ at output space, a binary cross entropy loss function given in Eq. \ref{eqn:4}, is used to train $D$. 
\begin{equation}
    L_d (P_{O}) = -\sum_{H, W} z ~log(D(P_{O}))+ (1-z)log(1 - D(P_{O}))
\label{eqn:4}
\end{equation}
where $z$ is an indicator variable and $z=0$ if the input segmentation output belongs to the target domain and $z=1$ if the input segmentation output belongs to the source domain dataset.
$D(P_{O}) \in \mathbb{R}^{H\times W\times 1}$ is the sigmoid output predicted by the discriminator network $D$.
Similarly, for latent space representation, the discriminator loss function is defined based on $P_{L}$ and is given by;
\begin{equation}
    L_d (P_{L}) = -\sum_{H, W} z ~log(D(P_{L})) + (1-z)log(1 - D(P_{L}))
\label{eqn:41}
\end{equation}
where $z$ is the same as defined in Eq. \ref{eqn:4} and $D(P_{L}) \in \mathbb{R}^{H\times W\times 1}$ is the sigmoid output predicted by the discriminator network $D$.

\textbf{Adaptation Network Training:} The adaptation network is composed of the segmentation network as a generator followed by the discriminator network as domain classifier. 
During adaptation, the generator network is trained using $L_{seg}(I_s, Y_s)$ (segmentation loss) given in Eq. \ref{eqn:3} and adversarial loss $L_{adv} (I_t)$ given in Eq. \ref{eqn:5} below:
\begin{equation}
L_{adv} (I_t) = -\sum_{H, W} log(D(G(I_t))^{(H;W;1)})
\label{eqn:5}
\end{equation}
Minimizing the adversarial loss $L_{adv} (I_t)$ minimizes the distribution gap between source and target domain images. It tries to update $G$ in such a way, that the segmentation probabilities or the latent space representation produced by $G$ for both source domain and target domain look similar. The optimization criterion given in Eq. \ref{eqn:2} is used to train the generator and discriminator network simultaneously. 



\subsection{Weakly-supervised Domain Adaptation}
The segmentation network in the mechanism described above fails to ensure the detection of the built-up regions when there are actually present.
The problem arises with the absence of a well defined structure in satellite imagery compared to other scene understanding datasets, e.g., urban driving scene segmentation.
These irregularities make the encoder and decoder of the segmentation network under-trained making the generic UDA methods (\cite{tsai2018learning, chen2017no}) less effective. (See Fig. \ref{img:401} (OSA, LTA) for qualitative comparison)

\begin{figure*}[t]
	\centering
	\includegraphics[width=4.8in]{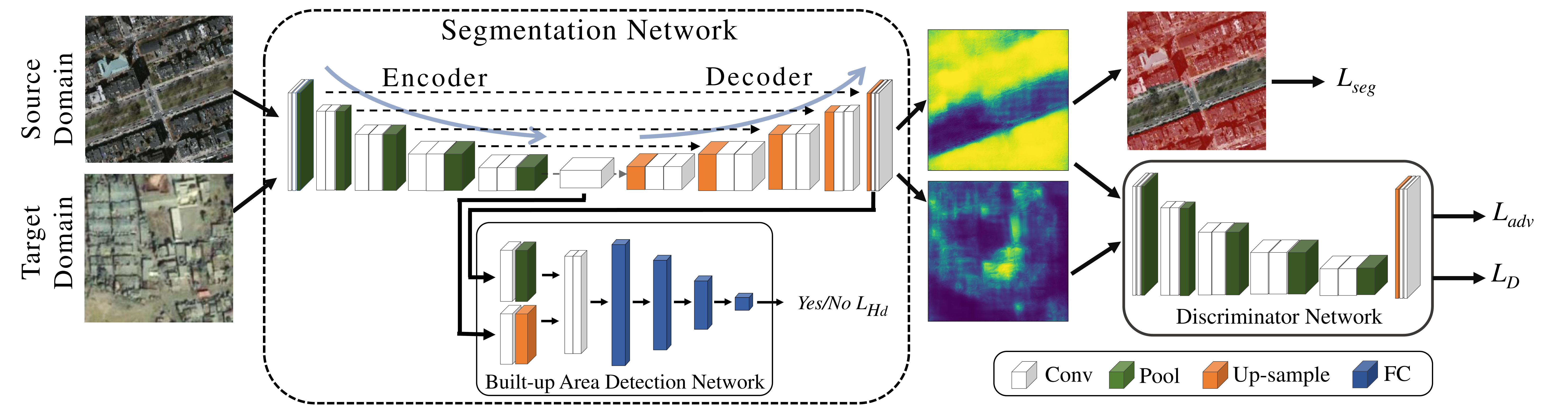}
	\caption{An overview of the proposed OS-WAN architecture. Like GAN based UDA methods for semantic segmentation, a fully convolutional discriminator network is trained using $L_D$. The segmentation network is optimized with $L_{adv}$ loss and $L_{seg}$ loss in output space, and an additional $L_{H_d}$ loss coming from built-up area detection network to both latent and output space.} 
	\label{img:3}
\end{figure*}

To mitigate these limitations, we have proposed a weakly supervised domain adaptation network (WAN) as shown in Fig. \ref{img:3}. It is difficult to collect dense labels for target domain datasets during UDA, however one can gather image level labels called weak-labels, which indicates the presence or absence of a certain object in the images. 
We use these weak-labels to help the adaptation method by nudging the source-trained network to understand how the objects in the target domain look like.
A small classification network trained for the image classification problem with these labels could be used to back-propagate the information and update the network.

\begin{figure*}[t]
	\centering
	\includegraphics[width=4.8in]{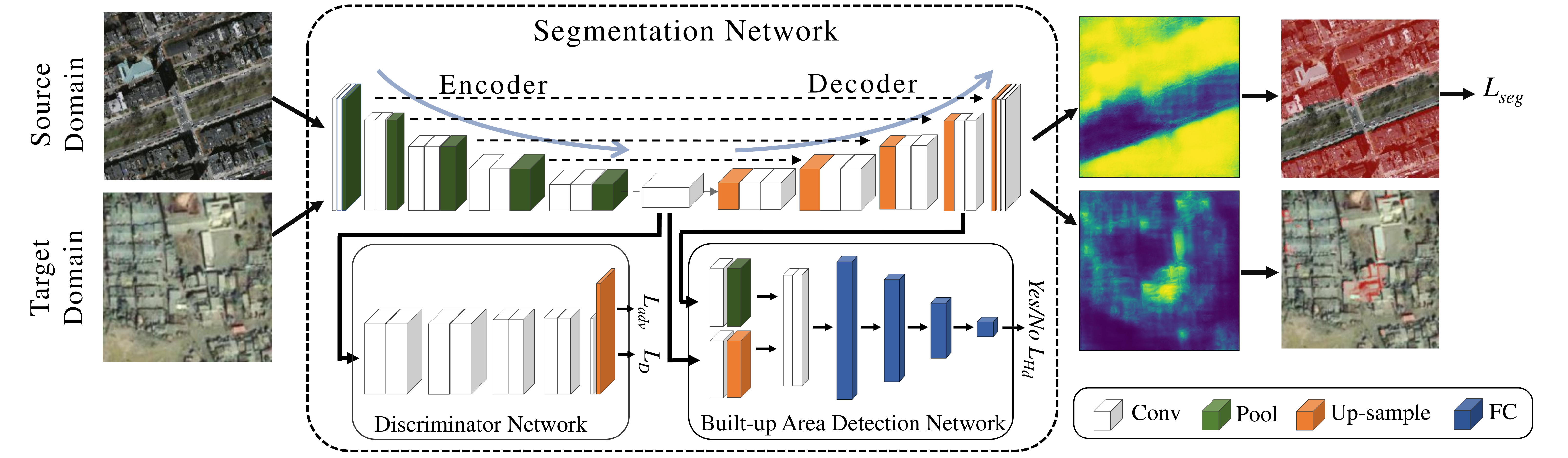}
	\caption{The figure demonstrates the proposed LT-WAN architecture for built-up regions segmentation. Like OS-WAN, a fully convolutional discriminator network is trained using $L_D$. The segmentation network is optimized with $L_{adv}$ loss from discriminator network in latent space, $L_{seg}$ loss in output space and $L_{H_d}$ loss in both latent and output space.} 
	\label{img:300}
\end{figure*}

For our problem, the weak-labels for the target domain just indicate if the image contains a built-up area or not, and the weakly supervised domain adaptation consists of training the \textit{built-up area detection network} alongside the UDA methods. This built-up area detection network once trained shall only give one binary value for the whole input image. 
The objective function for built-up area detection network is a binary cross entropy loss shown in  Eq. \ref{eqn:wl-7}.
\begin{equation}
\small
L_{H_d} (I_t) = - [(b_H)log(H_d(I_t)) + (1-b_H)log(1 - H_d(I_t))]
\label{eqn:wl-7}
\end{equation}
where $b_H$ is an indicator variable and $b_H=0$ if the image has no buildings or built-up regions, and $b_H=1$ if the input image has built-up regions. $H_d(I_t) \in \mathbb{R}^{1}$ is the binary output predicted by built-up area detection network $\mathrm{H_{d}}$ for an input target image $I_t$. 
For domain adaptation, $H_d$ makes its decision by using features computed from the model to be adapted.
The error in the classification while adaptation is back propagated to the network, thus forcing it to learn to capture the features necessary for the classification of objects (buildings in current case). 
To adapt the segmentation model
, a building (built-up regions) detection network is integrated with the segmentation network, as shown in Fig. \ref{img:3} and Fig. \ref{img:300}, and is trained using the available weak-labels for the target domain.

In this work, we have augmented the proposed built-up area detection network with both latent space adaptation (LTA) and output space adaptation (OSA) approaches which are described in detail below.



\subsubsection{Output Space Weakly-supervised Domain Adaptation}
\label{sec:os-wan}

The Output  space weakly-supervised domain adaptation network (OS-WAN), like previous adversarial domain adaptation methods, is composed of two major sub-networks, e.g., generator network $G$ and discriminator network $D$.
The U-Net segmentation network that has been previously trained to segment source images, is used as the generator $G$. 
The built-up area detection network, denoted as $H_d$, is integrated in $G$ by extracting features at both the encoder and decoder level as shown in Fig. \ref{img:3}. 
The segmentation network generates dense predictions while the built-up area detection network predicts the presence of built-up areas. The objective function for segmentation network is the same as defined in Eq. \ref{eqn:3}.
The overall cost function for the new generator network (segmentation network and the built-up area detection network) is the composition of both the above losses and is given by Eq. \ref{eqn:wl-8}
\begin{equation}
L_{gen} (I_s,Y_s, I_t) = L_{seg} (I_s, Y_s) + \alpha_{H_d} L_{H_d} (I_t)
\label{eqn:wl-8}
\end{equation}
where $\alpha_{H_d}$ is the weighing factor to control the effect of built-up area detection network loss.
Similar to segmentation network, the discriminator network in the OS-WAN approach is the same as described in Section \ref{sec:uda} for UDA. It is trained in a similar fashion using the loss described in Eq. \ref{eqn:4}.

In the proposed OS-WAN approach, the generator network $G$ is trained based on $L_{gen} (I_s, Y_s, I_t)$ given in Eq. \ref{eqn:wl-8} as well as adversarial loss $L_{adv} (I_t)$, Eq. \ref{eqn:5}. The adversarial loss minimizes the global distribution gap between source and target datasets while $L_{H_d}$ is used to help the generator to learn features that could be used to detect whether an input image has building(s) in it or not.
The $L_{seg}$ acts as a regularizer to remember the original task.
The combined loss function for OS-WAN is given in Eq. \ref{eqn:9}.
\begin{equation}
L_w{(I_s, I_t)} = L_{gen} (I_s, Y_s, I_t) + \lambda_{adv} L_{adv} (I_t)
\label{eqn:9}
\end{equation}
where $L_w{(I_s, I_t)}$ is the combined loss function, the subscript $w$ represents a weakly-supervised approach along with output space or latent space adversarial loss. $L_{gen} (I_s)$ is the generator loss and is the combination of segmentation and built-up area detection losses of the source and target domain respectively. Similarly, $L_{adv} (I_t)$ is the adversarial loss and $\lambda_{adv}$ is the scaling factor to limit the effect of adversarial loss.  
$L_w{(I_s, I_t)}$ given in Eq. \ref{eqn:9} can be expanded to individual components as shown in Eq. \ref{eqn:10}:
\begin{equation}
L_w{(I_s, I_t)} = L_{S} (I_s, Y_s) + \alpha_{H_d} L_{H_d} (I_t) + \lambda_{adv} L_{adv} (I_t)
\label{eqn:10}
\end{equation}
Finally, optimization equation $\underset{G}{min}  ~\underset{D}{max}  ~L_w{(I_s, I_t)} $ is used to minimize the  over generator and maximize it over discriminator network  


\subsubsection{Latent Space Weakly-supervised Domain Adaptation}
Output space domain adaptation has been shown to result in high accuracy while adapting to target domain if both domains consist of samples with structure, e.g., urban road scene segmentation.
However, it is observed that it is less effective when there is no well defined structure or any other spatial constraints available, e.g., aerial or satellite imagery datasets. In this section we exploit the adversarial domain adaptation in latent space integrated with the proposed weakly-supervised adaptation. 

The generator network $G$ of the latent space weakly-supervised domain adaptation (LT-WAN) approach is similar to that of the OS-WAN and is optimized using Eq. \ref{eqn:wl-8} for source domain segmentation and target domain building detection.
Similarly, the discriminator network $D$ in LT-WAN is a fully convolutional network applied at the latent space representations of the segmentation part of the generator network. 
Using these latent space representations defined as $P_L = G(I) \in \mathbb{R} ^{h\times w\times c}$, $D$ is trained with a binary cross entropy loss to discriminate the $P_L$ of source and target domains as given by Eq. \ref{eqn:4}.

In the proposed LT-WAN, the generator network $G$ during adaptation is trained based on $L_{gen} (I_s,Y_s, I_t)$ as well as adversarial loss $L_{adv} (I_t)$ given in Eq. \ref{eqn:9}. In LT-WAN, contrary to application of Eq. \ref{eqn:9} of OS-WAN, the adversarial loss is applied only at the encoder output of the segmentation part of the generator network (Fig. \ref{img:300}). The rest of the optimization process is the same as the OS-WAN described in Section \ref{sec:os-wan}.

\section{Experiments and Results} \label{Experiments}
In this section, we describe our experimental setup and obtained results in detail. 

\subsection{Experimental Setup}

\subsubsection{Datasets}
\label{sec:datasets}
Fig. \ref{img:2} visually presents the difference between our single source and multiple target domains datasets. Structures present in the Massachusetts and ISPRS-Potsdam images though look different from one another but are built in regular patterns compared to Rwanda and village Finder (VF) datasets which have irregular built-up areas. The datasets used to evaluate the proposed OS-WAN and LT-WAN are explained below.
\begin{table}[h]
	\centering
	
	\caption{Publicly available and the created datasets details before pre-processing. (m/p*: number of square meters per pixel, Patch Size*: original image size provided in dataset)}
	\scriptsize
	\begin{tabular}{|P{2.2cm}|P{1.1cm}|P{1.1cm}|P{1.1cm}|P{1.1cm}|P{1.5cm}|P{1.2cm}|}
		\hline
		Dataset& Total Images&  Training Images& Validation Images& Testing Images& Patch Size* &Resolution (m/p*)\\ \hline
		Massachusetts ~(\cite{MnihThesis}) &  151&  137& 4& 10& 1500 $\times$ 1500 &1.0\\ \hline
		Village Finder ~(\cite{murtaza2009villagefinder}) &  60&  25& 10& 25& 2400 $\times$ 2400 & 0.54\\ \hline
		ISPRS-Potsdam ~(\cite{ISPRS_potsdam}) &  38&  16& 2& 18& 6000 $\times$ 6000 & 0.05\\ \hline
		Rwanda (Ours) &  787&  300& 100& 387& 256 $\times$ 256&1.193\\ \hline
	\end{tabular}
	\label{table:1e}
\end{table}


\textbf{Massachusetts Buildings Dataset:} This dataset consists of 151 tiles of size $1500\times1500$, out of which 137 tiles are provided for training, 4 for validation and 10 for testing as shown in Table \ref{table:1e}. Even patches of size $500\times500$ are cropped and resized to $512\times512$ from each tile image. Each $512\times512$ patch is further divided into $256\times256$ patches as standard image size for this work. The dataset is cleaned by removing the empty patches (empty locations in the provided large image tiles). The dataset details after pre-processing are provided in  Table \ref{table:1}. The Massachusetts building dataset is a single source high resolution (HR) dataset covering (1.0 meter per pixel) a single geographical location augmented with similar built-up structures. 
  
\textbf{Village Finder Dataset:} The Village Finder (VF) dataset is created by collecting images from multiple locations of nucleated villages (\cite{murtaza2009villagefinder}). The proposed model adopts the structures of developed urban areas using the information provided by low resolution nucleated villages data. These are very high resolution (VHR) images covering 0.54 meters per pixel (see Table. \ref{table:1e}). Out of the provided 60 tiles of size $2400\times2400$, 25 tiles are used as training set, 10 as validation and the remaining 25 as testing set respectively. Each large tile is further divided into $25$ small tiles, resized to $512\times512$ and then divided into $256\times256$ images as a standard image size.  Table \ref{table:1} shows the pre-processed dataset information.  
  
\textbf{ISPRS 2D Semantic Labeling Potsdam:} ISPRS Potsdam dataset is the highest resolution publicly available dataset for house segmentation (VHR aerial imagery). The ground sampling distance of Potsdam's images is 5cm. Out of available 38 tiles of size $6000\times6000$, 16 tiles are used for training, 2 for validation and 18 for testing respectively as shown in  Table \ref{table:1e}. 

In order to bring uniformity across datasets, we choose an image size of $256\times256$ for segmentation task and extract patches of this size from the original dataset. In case of ISPRS-Potsdam, since the resolution is too high, we resize them too.
The $6000\times6000$ tiles are divided into 16 sub-images $1500\times1500$. To make the buildings cover a small area, each $1500\times1500$ is resized to $512\times512$. For segmentation network input size and to match memory limitations, each $512\times512$ sub-image is divided into 4 patches of size $256\times256$. The final number of training, validation and testing images are shown in  Table \ref{table:1}. This process overcomes the receptive field and memory issues and significantly improves the segmentation performance by incorporating enough context alongside buildings.
\begin{table}[h]
	\centering
	
	\caption{The pre-processed publicly available and the created datasets for domain adaptation. All the datasets are divided into fixed sized sub-images as described in Section \ref{sec:datasets}.}
	\scriptsize
	\begin{tabular}{|P{2.0cm}|P{1.8cm}|P{1.8cm}|P{1.8cm}|P{1.8cm}|}
		\hline
		Dataset & Total Images &  Training Images& Validation Images& Testing Images\\ \hline
		Massachusetts &  4788&  4284& 144& 360\\ \hline
		Village Finder &  6000&  2500& 1000& 2500\\ \hline
		ISPRS-Potsdam &  608&  256& 80& 272\\ \hline
		Rwanda (Ours) &  787&  300& 100& 387\\ \hline
	\end{tabular}
	\label{table:1}
\end{table}
  
\textbf{Rwanda Built-up Regions Dataset:} The proposed Rwanda built-up regions dataset is very much different in nature from previously available datasets. The varying structure size and formation, irregular patterns of construction, buildings in forests and deserts, and the existence of mud houses makes it very challenging. A total of 787 satellite images of size $256\times256$ are collected at high resolution (HR) of 1.193 meter per pixel (Table. \ref{table:1e}) and hand tagged for built-up region segmentation using an online tool Label-Box (\cite{labelBox}). The difference between Rwanda images and other existing datasets images is shown in Fig. \ref{img:2} which makes it very challenging for an adaptation algorithm. The variations in visual appearances and terrain are successfully captured by our proposed adaptation method as shown in Table. \ref{table:2}.

\subsubsection{Network Architectures}  \hfill \break
\textbf{Generator Network:} To achieve high quality built-up regions segmentation results, U-Net based encoder-decoder network is trained as a baseline segmentation model. 
Many of the neural network designs that have been proposed (\cite{treeSegNet, ronneberger2015u, yasrab2016scnet}) for the segmentation tasks, can be said to contain a basic design of the encoder-decoder network. 
Basically, as the name indicates, such a network consists of two sub-networks. The encoder, in general, maps the input to a low dimensional latent space. In the case of image segmentation, the encoder accumulates features such that spatial information is represented in compressed form. 
Decoder consists of up-sampling steps and CNN layers, such that the final output is of the same spatial size as input and for each pixel we have the probability of which semantic segment it belongs to.
In this work, we have used medium depth, e.g., 32 filters instead of 64 filters in the first layer and doubling in each next layer of the encoder in the U-Net architecture due to resource limitations. 
In UDA approach, the generator consists of segmentation network only, i.e., U-Net architecture for foreground background segmentation.
However, during weakly-supervised domain adaptation setup, the generator is composed of a segmentation network integrated with a built-up area detection network as described in Section \ref{sec:os-wan}.
built-up area detection network is a binary classification network built over the encoder and decoder of the segmentation network. The latent space representation of the U-Net is up-scaled twice and the last feature map of the decoder is down-scaled twice using bi-linear interpolation and are concatenated depth-wise in a single layer as shown in Fig. \ref{img:3} and \ref{img:300}. The concatenated output is followed by two convolutional layers and three fully connected layers with relu activation except the last layer, where sigmoid is applied.

\textbf{Discriminator Network:}
The discriminator $D$ is a fully convolutional network inspired from (\cite{tsai2018learning}). 
For OSA (both unsupervised and weakly-supervised) the discriminator network is composed of 5 convolutional layers with depths of {64, 128, 256, 512 and 1} respectively while the kernel size of $4 \times 4$ pixels with stride 2 in each layer. 
For latent space adaptation, LTA, (both unsupervised and weakly-supervised) the discriminator network have 5 convolutional blocks having {256, 256, 128, 64 and 1} respectively. The kernel size in this case is $3 \times 3$ with a stride of 1. 
In both cases, each convolutional layer is followed by a LeakyRelu activation (\cite{maas2013rectifier}) parameterized by 0.2 except the last layer where sigmoid is applied. An upsampling layer is added in-order to match the output size to the input size. No batch normalization and pooling layers are used. 

\subsubsection{Implementation Details}

To have a baseline, the segmentation network is trained in a fully supervised manner on source domain datasets by optimizing $L_{seg}$ in Eq. \ref{eqn:3}. For domain adaptation, the images $I_s$ from source domain and $I_t$ from target domain are passed through the trained generator (segmentation) network to generate $G(I_s)$ and $G(I_t)$ respectively. These representations (either at latent space or output space) are used to train the discriminator network by optimizing $L_d$ and at the same time $L_{adv}$ is back-propagated to update the generator network. At the same time, the $L_{H_d}$ for built-up area detection network is back propagated to update the segmentation network as well. The process of training the discriminator and generator is repeated for mini-batches until the network is stabilized. 

We implement both the domain adaptation approaches (OS-WAN and LT-WAN) in Keras toolbox (\cite{chollet2015keras}) with tensorflow (\cite{abadi2016tensorflow}) backend on a single Core-i5 machine with 32GB RAM and a GTX 1080 GPU with 8GB of memory. We used Adam optimizer (\cite{kingma2014adam}) with initial learning rate of 0.001 for segmentation network, 0.0001 and 0.00001 for discriminator networks of two setups (output and latent space) and 1e-6 for adversarial learning respectively. The momentum is set as 0.9 and 0.99 while the weight decay is set 1e-6 respectively. The hyperparameter $\lambda_{adv}$ is set to 0.1 for OS-WAN and 0.01 for LT-WAN. Similarly, $\alpha_{H_d}$ is set to 0.1 for both approaches. 

\subsection{Experimental Results}

In this section, a quantitative and qualitative evaluation of the proposed schemes for domain adaptation of aerial and satellite imagery is provided. As described in Section \ref{sec:datasets} and Table \ref{table:1e}, three publicly available datasets and a collected dataset are processed for built-up region segmentation. Multiple experiments  with defined network configurations are performed. In all the experiments, standard evaluation metrics, e.g., intersection over union (IoU) and F1-score are used (\cite{demir2018deepglobe}). 
Similarly, in all major experiments reported in Table. \ref{table:2}, Massachusetts buildings dataset is used as source domain while all other datasets are used as target domains. To have a comprehensive validation of the proposed dataset, we have performed experiments with the remaining three datasets as source domain dataset one by one (see Table. \ref{table:comparisonAll}).

\begin{table*}[!htb]
	\centering
	\caption{\small{Performance comparison of different strategies when adapted from  Massachusetts dataset to Village Finder, ISPRS-Potsdam and Rwanda datasets respectively. The upper bound is the maximum performance the model can achieve as defined in Sec. \ref{sec:discussion}}}
	\label{table:2}
	\scriptsize
	\begin{tabular}{|P{1.7cm}|P{1.2cm}|P{1.3cm}|P{1.1cm}|P{1.2cm}|P{1.1cm}|P{1.2cm}|}
		\hline
		Source $\rightarrow$ Target& \multicolumn{2}{l|}{Massachusetts $\rightarrow$ Village Finder}& \multicolumn{2}{l|}{Massachusetts $\rightarrow$ Potsdam}& \multicolumn{2}{l|}{Massachusetts $\rightarrow$ Rwanda}\\ \hline
		Method&   IoU& F1-Score&   IoU& F1-Score&   IoU& F1-Score \\ \hline

		Baseline(U-Net)&  24.41& 39.25&   19.18& 32.18&   6.37& 11.98\\ \hline
		OSA ~(\cite{tsai2018learning}) &   42.33& 59.47&  33.96& 50.71&  18.72& 31.54\\ 
		LTA ~(\cite{chen2017no}) &   46.01& 63.02&  38.83& 55.94&  24.11& 38.85\\ 
		OS-WAN (ours)&   47.64& 64.53& 40.10& 57.25& 32.10& 48.60\\ 
		LT-WAN (ours)&   \textbf{51.34}& \textbf{67.84}&  \textbf{45.38}& \textbf{62.43}&   \textbf{36.66}& \textbf{53.65}\\ 
		\hline
		Upper Bound&   67.94& 80.91&   58.59& 73.89&   72.21& 83.86\\ \hline
		
	\end{tabular}
\end{table*}

\subsubsection{Massachusetts $\rightarrow$ Village Finder}
In Table \ref{table:2}, we report the built-up region segmentation performance in terms of F-1 score and IoU over the Village Finder testing set. The proposed weakly-supervised domain adaptation methods outperform the baseline and existing state-of-the-art algorithms. 
As described, U-Net is used as base-line segmentation architecture. The segmentation capability of U-Net is increased during adaptation with the integration of proposed built-up area detection network.
The composite model is now able to correctly segment built-up regions by maximizing the true detections and minimizing the false segmentation as shown in Table \ref{table:2}. 

During adaptation, the built-up area detection network in the composite model forces the encoder and decoder of the segmentation network to extract features of built-up region existence in an image, which eventually leads to better adaptation and segmentation. The loss of the built-up area detection module is weighed by $\alpha_{H_d}$, chosen as 0.1. The results of both target and source domains before and after adaptation are given in Table \ref{table:2} and \ref{table:3} respectively. 

\begin{figure}[H]
	\centering
	\includegraphics[width=\textwidth]{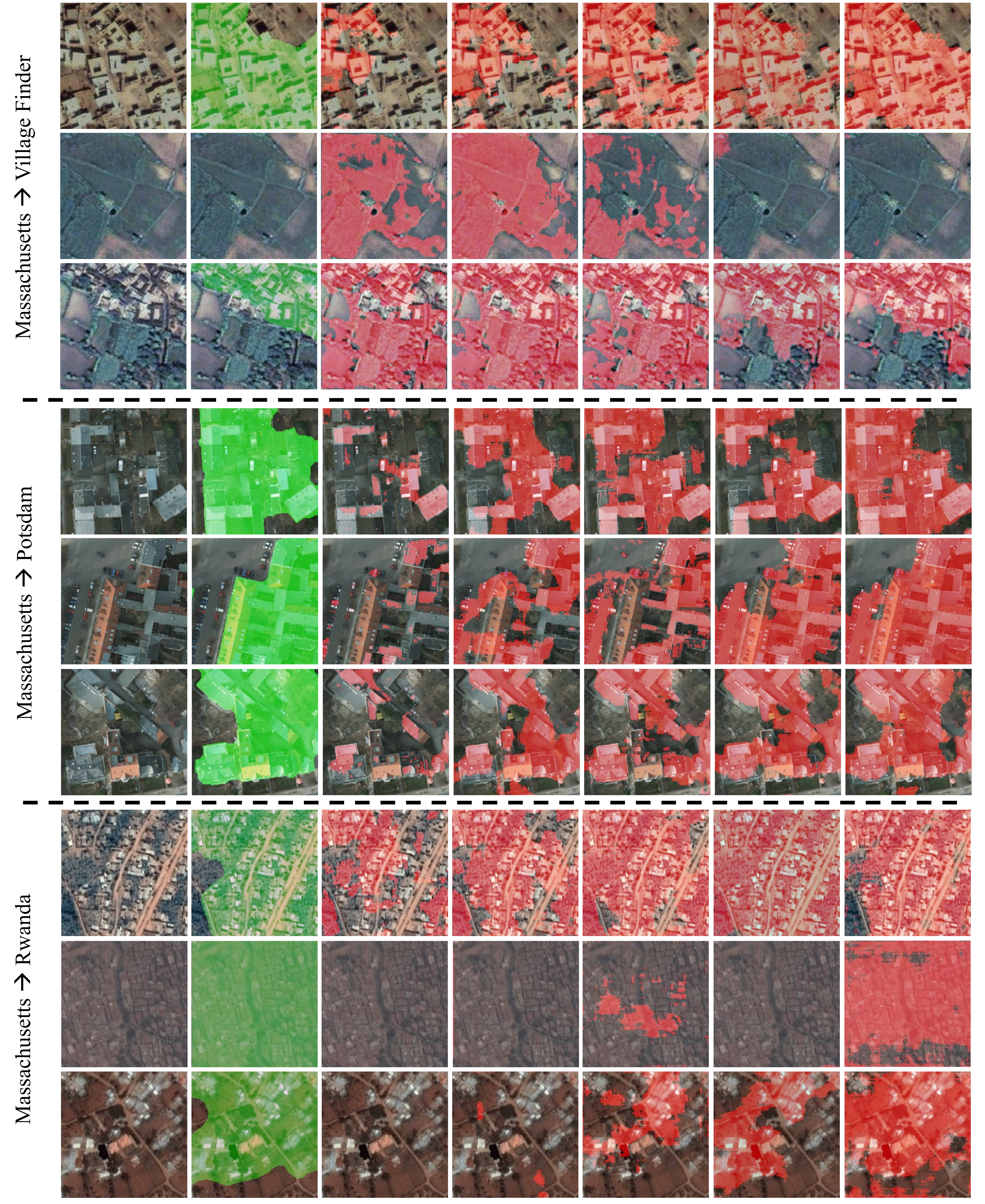}
	\scriptsize
	\begin{tabular}{P{0.1cm}P{1.3cm}P{1.35cm}P{1.25cm}P{1.0cm}P{1.1cm}P{1.4cm}P{1.2cm}}
		&Target Image &  Ground Truth& Source Only&  OSA & LTA &  OS-WAN (ours)& LT-WAN (ours) \\ 
	\end{tabular}
	\caption{Output of built-up area segmentation when adapted from Massachusetts to Village-Finder, Potsdam and Rwanda datasets respectively. Columns from left to right represent Target Image, Ground Truth, Source Only, OSA: output space adaptation (\cite{tsai2018learning}), LTA: latent space adaptation (\cite{chen2017no, benjdira2019unsupervised}), OS-WAN and LT-WAN respectively. The proposed LT-WAN and OS-WAN mitigates the false segmentation posed by OSA and LSA and improves the true positive rate significantly.}
	\label{img:401}
\end{figure}

\begin{figure}[h]
	\centering
	\includegraphics[width=\textwidth]{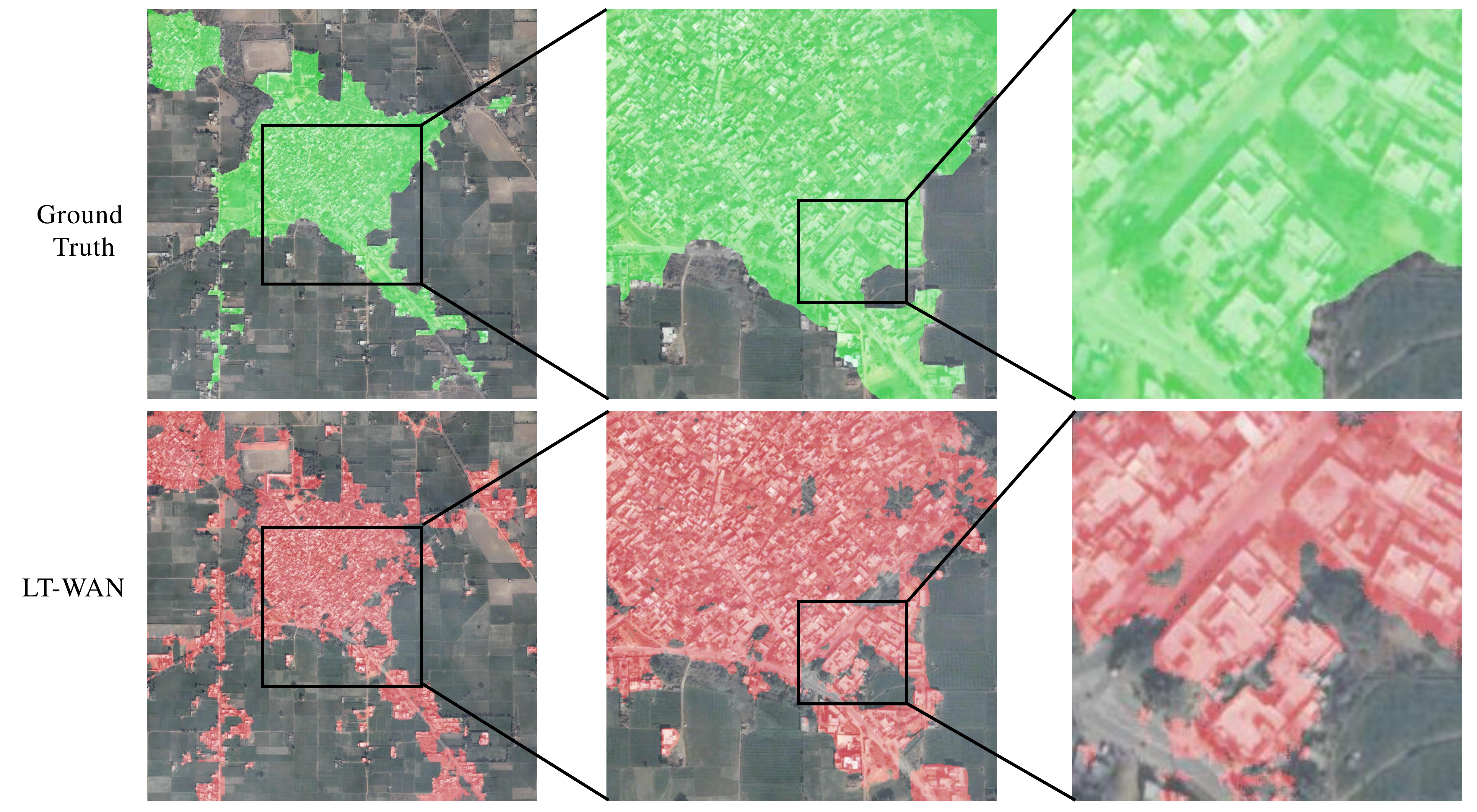}
	\scriptsize
	\caption{Built-up area segmentation of a whole nucleated village when adapted from Massachusetts to Village-Finder. (best viewed in color)}
	\label{img:large_res_vf}
\end{figure}

Qualitative results of the proposed OS-WAN and LT-WAN algorithms are shown in Fig. \ref{img:401}, comparing them to baseline built-up region segmentation network and other UDA methods over Village Finder dataset. 
OS-WAN improves the segmentation true positive rate and successfully mitigates the false positive segmentation. The LT-WAN shows further refined results with better separation boundaries and built-up areas. 
As discussed in Section \ref{sec:datasets}, Village Finder is a multi-distribution dataset in nature (nucleated villages images), so the gain due to domain adaptation is limited compared to other datasets.
However, OS-WAN architecture still achieves a significant gain of 3.4\%, 11.2\% and 48.8\% in IoU and 2.3\%, 7.8\% and 39.2\% increase in F1-Score over the latent space, output space unsupervised domain adaptation and baseline model respectively. 
Similarly, the proposed LT-WAN algorithm outperforms the OS-WAN, latent space and output space UDA methods and baseline model by 7.2\%, 10.5\%, 17.6\% and 52.5\% in IoU and 4.9\%, 7.1\%, 12.3\% and 42.1\% in F1-Score respectively.
Fig. \ref{img:large_res_vf} shows the built-up region segmentation output for a whole nucleated village and its closer versions from the Village Finder test set. 
The results indicate that LT-WAN has better performance while domain adaptation in challenging scenarios. 

\vspace{-0.1cm}
\vspace{-0.1cm}

\subsubsection{Massachusetts $\rightarrow$ Potsdam}
Table \ref{table:2} shows the results of the Potsdam testing set. It is observed that the Potsdam dataset contains images of well planned development which makes it viable for the segmentation network to adapt easily to this dataset.
The only difference is that Potsdam has large size houses and very high resolution imagery compared to the source Massachusetts dataset making it difficult for a simple segmentation network to capture the global view. The proposed weakly supervised methods achieve state-of-the-art performance over the baseline and other existing methods. The OS-WAN architecture achieves 3.2\%, 15.3\% and 50.32\% gain in IoU and +2.3\%, 11.4\% and 43.8\% increase in F1-Score over the latent space, output space UDA methods and baseline model respectively. 
Similarly, the proposed LT-WAN architecture achieves 11.6\%, 14.4\%, 25.2\% and 57.73\% IoU increase and 8.3\%, 10.4\%,  18.8\% and 48.5\% gain in F1-Score over the OS-WAN, latent space and output space UDA methods and baseline model respectively.
Fig. \ref{img:401} shows the qualitative results of the proposed OS-WAN and LT-WAN for Massachusetts to Potsdam adaptation. Despite the variation in building sizes and structures, the proposed weakly-supervised approaches capture the target dataset far better compared to OSA and LTA.

\vspace{-0.15cm}
\subsubsection{Massachusetts $\rightarrow$ Rwanda}
As discussed in Section \ref{Experiments} and shown in Fig. \ref{img:2} in detail, the developed Rwanda dataset poses some unique features like mud houses, small built-up regions in forests and deserts, irregular patterns and tin rooftops. These features make it more realistic and more difficult for segmentation algorithms to adapt to such datasets. Table \ref{table:2} shows results of the Rwanda test set before and after adaptation. The proposed weakly supervised methods show superior performance compared to base-line segmentation models and other state-of-the-art UDA algorithms. Using OS-WAN, we achieve 24.9\%, 41.7\% and 80.2\% gain in IoU and 20.1\%, 35.1\% and 75.4\% increase in F1-Score over the latent and output space UDA methods and the baseline model respectively.
Similarly, the proposed LT-WAN architecture achieves 12.4\%, 34.2\%, 48.9\% and 82.6\% IoU increase and 9.4\%, 27.6\%, 41.2\% and 78.2\% increase in F1-score over the OS-WAN, latent and output space UDA methods and baseline model respectively. 
The qualitative results of Massachusetts to Rwanda adaptation described in Fig. \ref{img:401} show the importance of weak supervision compared to baseline U-Net, OSA and LTA approaches. LT-WAN significantly improves the true positive and false negative rates producing better segmentation with finer separation details.

\subsubsection{Village Finder, Potsdam and Rwanda as Source Domain Datasets}
To have a comprehensive evaluation of the proposed weakly-supervised OS-WAN and LS-WAN, we have performed experiments with Village Finder, Potsdam and Rwanda as source domain dataset one by one. Due to limitations in computational resources, for each source dataset, we choose the target dataset for which either OS-WAN or LT-WAN has the smallest gain. For that source-target pair we report results for all other algorithms. While for the rest we report the proposed LT-WAN and OS-WAN only (Table. \ref{table:comparisonAll}).

\begin{table*}[!htb]    
	\centering
	\caption{\small{Evaluation of the proposed weakly-supervised approaches compared to baseline U-Net.  Source $\rightarrow$ Target means that the segmentation model is trained on source domain and adapted to target domain, i.e., in case of Potsdam $\rightarrow$ Massachusetts, the segmentation model is trained on Potsdam dataset and then adapted to Massachusetts dataset.
	}}
	\label{table:comparisonAll}
	\scriptsize
    \begin{tabular}{|P{1.7cm}|P{1.2cm}|P{1.3cm}|P{1.1cm}|P{1.2cm}|P{1.1cm}|P{1.2cm}|}
		\hline
		Source $\rightarrow$ Target& \multicolumn{2}{l|}{Potsdam $\rightarrow$ Massachusetts}& \multicolumn{2}{l|}{Potsdam $\rightarrow$ Village Finder}& \multicolumn{2}{l|}{Potsdam $\rightarrow$ Rwanda}\\ \hline
		Method&   IoU& F1-Score&   IoU& F1-Score&   IoU& F1-Score \\ \hline
		Baseline(U-Net)&  41.3& 58.4&   40.0& 57.2&   18.3& 31.0\\ 
		OSA ~(\cite{tsai2018learning})& -&-&40.8&58.0&-&-\\
		LTA ~(\cite{chen2017no}) & -& -&41.0&58.1&-&-\\
		OS-WAN (ours)&   66.0&  79.5& 44.2& 61.3& 31.6& 48.0\\ 
		LT-WAN (ours)&   \textbf{69.2}& \textbf{81.8}&  \textbf{44.5}& \textbf{61.6}&   \textbf{39.8}& \textbf{56.9}\\ 
		\hline
		
		Source $\rightarrow$ Target& \multicolumn{2}{l|}{Village Finder $\rightarrow$ Massachusetts}& \multicolumn{2}{l|}{Village Finder $\rightarrow$ Potsdam}& \multicolumn{2}{l|}{Village Finder $\rightarrow$ Rwanda}\\ \hline
		Baseline(U-Net)&  68.9& 81.6&   8.7& 16.0&   22.4& 36.7\\ 
		OSA ~(\cite{tsai2018learning})& 69.0& 81.7&-&-&-&-\\
		LTA ~(\cite{chen2017no}) & 71.9& 82.3&-&-&-&-\\
		OS-WAN (ours)&   72.7&  84.2& 26.0& 41.2& 39.8& 56.9\\ 
		LT-WAN (ours)&   \textbf{77.6}& \textbf{87.4}&  \textbf{28.8}& \textbf{44.7}&   \textbf{45.0}& \textbf{62.1}\\ 
		\hline
		
		Source $\rightarrow$ Target& \multicolumn{2}{l|}{Rwanda $\rightarrow$ Massachusetts}& \multicolumn{2}{l|}{Rwanda $\rightarrow$ Village Finder}& \multicolumn{2}{l|}{Rwanda $\rightarrow$ Potsdam}\\ \hline
		Baseline(U-Net)&  41.9& 59.1&   27.2& 42.8&   29.6& 45.7\\ 
		OSA ~(\cite{tsai2018learning})& -& -&-&-&36.1&53.0\\
		LTA ~(\cite{chen2017no}) & -& -&-&-&36.6&53.6\\
		OS-WAN (ours)&   69.3&  81.9& 47.6& 64.5& 39.6& 56.8\\ 
		LT-WAN (ours)&   \textbf{72.0}& \textbf{83.7}&  \textbf{51.7}& \textbf{68.2}&   \textbf{40.2}& \textbf{57.3}\\ 
		\hline
		
	\end{tabular}
\end{table*}
\begin{itemize}
    \item Potsdam as Source Domain: Table. \ref{table:comparisonAll} describes the detailed quantitative results of Potsdam's trained model adapted to Massachusetts, Village Finder and Rwanda using OS-WAN and LT-WAN. Potsdam is a high resolution imagery dataset with large sized buildings constructed in a defined pattern. While adapting Potsdam's trained model to Massachusetts, the OS-WAN and LT-WAN improves the IoU performance by 37.4\% and 40.3\% respectively. Similarly, when adapted to Village Finder and Rwanda, the proposed OS-WAN and LT-WAN outperform the baseline models with significant margins.
    To have fair comparison, we have applied OSA and LTA for Potsdam to Village Finder adaptation. As shown in Table. \ref{table:comparisonAll} (Potsdam $\rightarrow$ Village Finder), the improvement is less compared to Rwanda and Massachusetts adaptation, depicting this a hard target domain. The OS-WAN and LT-WAN outperforms OSA and LTA with significant margins indicating the role of weak-supervision. 
    
    \item Village Finder as Source Domain: Village Finder dataset has a variety in illumination and built-up structures. However, it does not have built-up regions present in Potsdam and Rwanda. Table. \ref{table:comparisonAll} shows the quantitative results of Village Finder's trained model adapted to Massachusetts, Potsdam and Rwanda datasets. The proposed OW-WAN and LT-WAN significantly improves the IoU and F1-score performances compared to the baseline model. To have a fair comparison with the existing algorithms, we have applied OSA and LSA for the hard case of Village Finder to Massachusetts adaptation when leveraged Village Finder as source domain. 
    
    \item Rwanda as Source Domain: Rwanda dataset is diverse in nature covering a variety of built-up structures. Table. \ref{table:comparisonAll} shows the detailed results of Rwanda's trained model adapted to Massachusetts, Potsdam and Village Finder datasets. The proposed OW-WAN and LT-WAN significantly improves the IoU and F1-score performances compared to the baseline model. Similar to Village Finder and Potsdam as source we have repeated all experiments for the hard case of Rwanda to Potsdam adaptation. As shown in Table. \ref{table:comparisonAll} (Rwanda $\rightarrow$ Potsdam), the proposed OS-WAN and LT-WAN outperforms the existing OSA and LTA with significant margins. Compared to baseline, OSA and LTA, the weak supervision significantly improves the performance.
\end{itemize}
\begin{table*}[!htb]
	\centering
	\caption{\small{Comparison of segmentation performance of Weakly-supervised segmentation method presented in (\cite{zhou2016learning_cam}) with proposed LT-WAN. This Weakly-supervised approach generates class activation maps (CAM) based on global average pooling (GAP) using a classification network. The final activations are thresholded at 0.5 to get segmentation output.
	}}
	\label{table:cam_}
	\scriptsize
	\begin{tabular}{|c|c|c|c|c|}
		\hline
		Method& \multicolumn{2}{c|}{(CAM~\cite{zhou2016learning_cam})}& \multicolumn{2}{c|}{LT-WAN (ours)} \\ \hline
		Dataset&   IoU& F1-Score &   IoU& F1-Score \\ \hline
		Village Finder~(\cite{murtaza2009villagefinder})&  14.5& 25.3&   51.34& 67.84\\ 
		Potsdam~(\cite{ISPRS_potsdam})&   20.2&  33.6&  45.38& 62.43\\ 
		Rwanda~(Ours)&   26.5& 42 &   36.66& 53.65\\ 
		\hline
	\end{tabular}
\end{table*}
\subsubsection{Weakly-supervised Segmentation}
In order to compare how weakly-supervised semantic segmentation performs against adaptation, we employed class activation maps (CAM), presented in (\cite{zhou2016learning_cam}). 
Table. \ref{table:cam_} shows the CAM based weakly-supervised segmentation results on Village Finder, Rwanda and Potsdam datasets. 
To make a fair comparison, we have used the encoding part of the U-Net as a backbone model and CAM layers were added over that. 
This technique however was only able to segment the prominent regions in an image patch. 
As indicated in Table. \ref{table:cam_}, segmentation results are not accurate compared to LT-WAN, which is reasonable since the structure does not have the skip layers that allow U-Net like architecture to make the fine-grain segmentation. Also the use of global average pooling leading to classification, complete and accurate segmentation is not required.   

\begin{table}[h]
	\centering
	\caption{Fine-tuning vs Domain adaptation: For both OS-WAN and LT-WAN, the results on the source dataset do not deteriorate compared to fine-tuning where a drastic reduction is noted.  
	}
	\label{table:3}
	\scriptsize
	\begin{tabular}{|P{3.2cm}|P{0.8cm}P{1.0cm}|P{0.8cm}P{1.0cm}|P{0.8cm}P{1.0cm}|}
		\hlineB{1.5}
		Process& \multicolumn{2}{c}{OS-WAN}& \multicolumn{2}{c}{LT-WAN}& \multicolumn{2}{c|}{Fine-tuning}\\ \hline
		Source $\rightarrow$ Target &  IoU& F1-Score&  IoU& F1-Score&  IoU& F1-Score \\ \hlineB{1.5}
	    Massachusetts Baseline &  84.36& 91.52&  84.36& 91.52 &  84.36& 91.52\\ \hlineB{1.5}
        Massachusetts $\rightarrow$ VillageFinder & 84.15& 91.39& 84.16& 91.40& 75.08& 85.76\\ 
        Massachusetts $\rightarrow$ Potsdam & 84.64& 91.68& 84.47& 91.57& 80.43& 89.15\\ 
        Massachusetts $\rightarrow$ Rwanda & 83.88& 91.23& 80.17& 88.99& 62.64& 77.03\\ \hline
	\end{tabular}
\end{table}
\subsection{Discussion}
\label{sec:discussion}
As shown in Fig. \ref{img:2}, the target domains have a variety of differences from source domain which makes it a difficult job for domain adaptation methods. Table \ref{table:2} shows the performance on the three target datasets for existing UDA methods and proposed weakly-supervised setups. Integrating $L_{H_d}$ with $L_{Adv}$, the proposed LT-WAN model achieves a gain of 10.5 - 34.2\% in IoU and 7.1 - 27.6 \% in F1 score compared to state-of-the-art domain adaptation algorithms. This shows that weakly-supervised loss in latent space and output space helps to improve the overall performance of the segmentation model significantly. Massachusetts and Potsdam datasets represent well structured regions but are very different from each other as shown in Fig. \ref{img:2}. The houses at the Potsdam are of large size which makes it difficult for baseline network trained on small houses in Massachusetts to segment. Besides these differences, the proposed adaptation network improves the performance to an observable difference as shown in Table \ref{table:2}. Similarly, the proposed OS-WAN and LT-WAN provide a considerable gain over Rwanda and Village Finder datasets where intra-dataset variations exist and can't be captured by a single unsupervised adaptation strategy. 

To define an upper limit (bound) for built-up region segmentation, the source domain trained baseline model is further fine-tuned on training sets of each target dataset respectively.  The test set performances for all the three target datasets are provided as "Upper Bound" in Table \ref{table:2}. Fine-tuning a trained model to a new target dataset have serious concerns of fully annotated data requirement and an observable drop in the performance of source domain dataset as shown in Table \ref{table:3}. The Massachusetts dataset observed 4.7 - 25.8\% drop in IoU and 2.6 - 15.83\% drop in F1-score during fine-tuning over provided target datasets. 
During the proposed domain adaptation process the performance of the original dataset is almost preserved as shown in Table \ref{table:3}. The Massachusetts dataset observed 0.57 - 4.9\% drop in IoU and 0.31 - 2.8\% drop in F1-score which is negligible compared to the fine-tuning approach as shown in Table \ref{table:3}. 

During OS-WAN it is observed that by integrating built-up area detection network with domain adaptation, the network converges quickly compared to simple adversarial adaptation.
The reason behind is, adversarial loss is back propagated through decoder, while the part of the built-up area detection loss is directly fed back to the encoder which makes the network to converge faster. 
While in case of LT-WAN, the adaptation process should be carried out slowly (with low learning rate and limited adversarial loss) in-order to keep the model stable. As the discriminator is applied at the encoder of the segmentation network, which makes it difficult for encoder to handle excess gradients, e.g., built-up area detection loss and adversarial loss with high learning rates. 
This strategy introduces an increase in adaptation time, but delivers better segmentation results compared to OS-WAN. 
Another advantage of LT-WAN is, it tries to match the distributions of source and target domain at feature level instead of output level. This helps the adaptation process for datasets where images lack scene structure like aerial and satellite imagery and hence produces better performance.

\section{Conclusion}

In this paper we have proposed a weakly-supervised domain adaptation network (WAN) for semantic segmentation of built-up regions which alleviates cross domain discrimination across diverse satellite and aerial imagery datasets.
We have followed a joint adversarial approach to transfer the target data distribution (structured output and latent space representation) of the segmentation network using a discriminator-generator pair.
To overcome the challenges posed by the aerial and satellite imagery such as large diversity and variance from the source domain and missing structure context mostly exploited (directly or indirectly) by previous adaptation methods, we introduce weakly supervised domain adaptation of the latent space as well as the output space.
Adaptation at high-dimensional latent space is improved by introducing a classification loss on the image level labels. 
This weakly-supervised loss forces latent space to adapt enough to capture information vital for the built-up area detection, before being sent to the decoder part.  
Weak labels allow the algorithm to adapt even for images where the base algorithm has near complete failure, in contrast to previous solutions that fail to adapt properly for such cases in general.
Furthermore, we have created a challenging Rwanda built-up regions segmentation dataset. 
The dataset captures a diverse set of terrains under multiple weather conditions. The Rwanda dataset also contains multiple kinds of built structures appearing to be concrete houses and mud buildings with tin and colored rooftops. 
To evaluate the proposed OS-WAN and LT-WAN, the segmentation model trained on Massachusetts dataset is adapted to Village Finder dataset, Potsdam dataset and the proposed Rwanda dataset with substantial improvement (11.6\%-52\% in IoU) from baseline and state-of-the-art unsupervised architectures.
For completeness the results on all source and target pairs is also computed and our approach outperforms the existing approaches.
The future directions are to extend this approach to other segmentation tasks and challenges in satellite and aerial imagery analysis.


\bibliography{arXiv}
\end{document}